%% file: root.tex
\documentclass[conference]{IEEEconf}
\IEEEoverridecommandlockouts    % to override locked commands
\usepackage{multirow}
\usepackage{lipsum}
\usepackage{hyperref}
\usepackage{xurl}
\usepackage{amsmath}
\usepackage{amssymb}
\usepackage{booktabs} % For professional lines
\usepackage{makecell}  % For the diagonal split cell
\usepackage{multirow}
\usepackage{graphbox}
\usepackage{subcaption}
\usepackage[ruled,vlined]{algorithm2e}
\usepackage{graphicx}
\usepackage{xcolor}
\usepackage{cuted}
\graphicspath{{./Figures/}}
\newcommand{\osdar}{OSDaR23}

\newcommand{\seqthree}{\texttt{3\_fire\_site\_3.1}}
\newcommand{\seqfive}{\texttt{5\_station\_bergedorf\_5.1}}
\newcommand{\seqsix}{\texttt{6\_station\_klein\_flottbek\_6.2}}
\newcommand{\rgbcenter}{\texttt{rgb\_center}}
\newcommand{\rgbhirescenter}{\texttt{rgb\_highres\_center}}
\newcommand{\URL}{\url{https://syndra.retis.santannapisa.it/osdarar.html}}
\title{\LARGE \bf OSDaR-AR: Enhancing Railway Perception Datasets via Multi-modal Augmented Reality}

\author{
 	\parbox{\textwidth}{%
 		\centering
 		Federico Nesti$^{1,2}$, Gianluca D'Amico$^{1,2}$, Mauro Marinoni$^{1,2}$, Giorgio Buttazzo$^{1}$%
 	}%
 	\thanks{$^{1}$Department of Excellence in Robotics \& AI, Scuola Superiore Sant'Anna, Pisa, Italy
 		{\tt\footnotesize<name>.<surname>@santannapisa.it}}%
 	\thanks{$^{2}$Simulatrix MV srl, Pisa, Italy}
}

\begin{document}
	
	%\maketitle
    % --- TEASER FIGURE LOGIC ---
\maketitle

%\vspace{-10mm}
\begin{strip}
    \centering
    %\vspace{-10mm} % Adjust spacing as needed
    \begin{minipage}{0.33\linewidth}
        \centering
        \vspace{-10mm}
        \includegraphics[width=\textwidth]{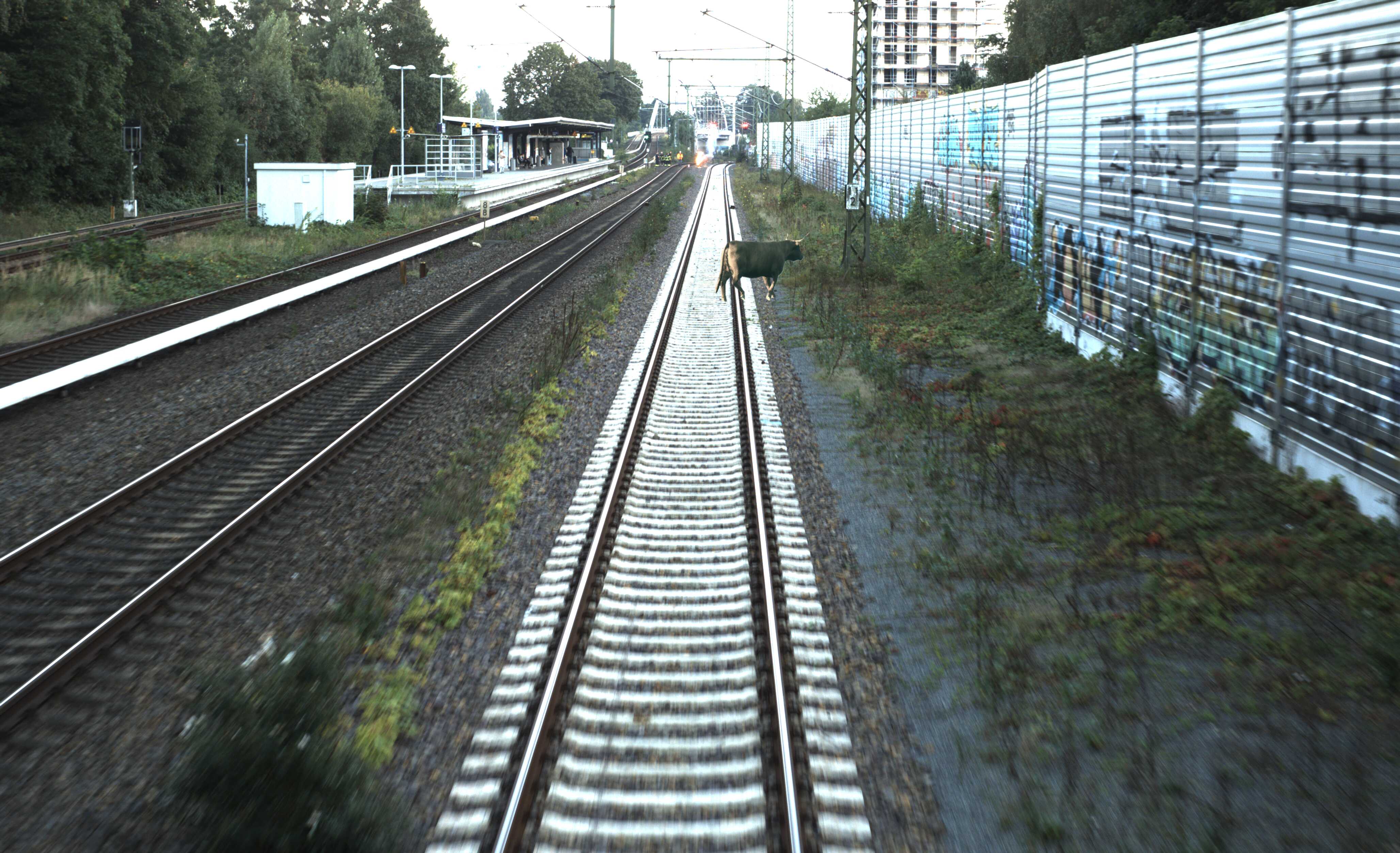}
        \\ \small (a)
    \end{minipage}
    \hfill
    \begin{minipage}{0.28\linewidth}
        \centering
        \vspace{-10mm}
        \includegraphics[width=\textwidth]{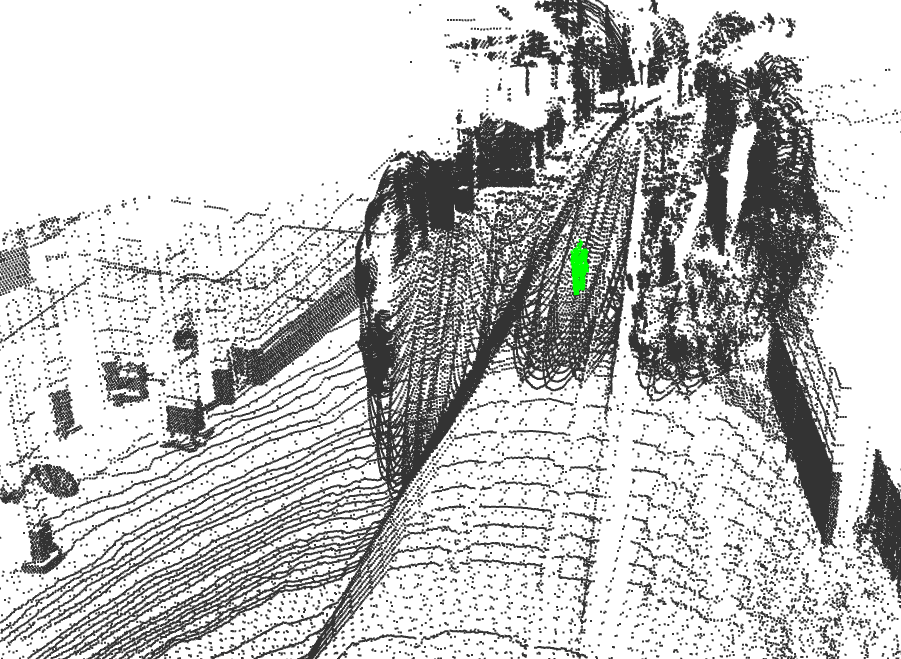}
        \\ \small (b)
    \end{minipage}
    \hfill
    \begin{minipage}{0.33\linewidth}
        \centering
        \vspace{-10mm}
        \includegraphics[width=\textwidth]{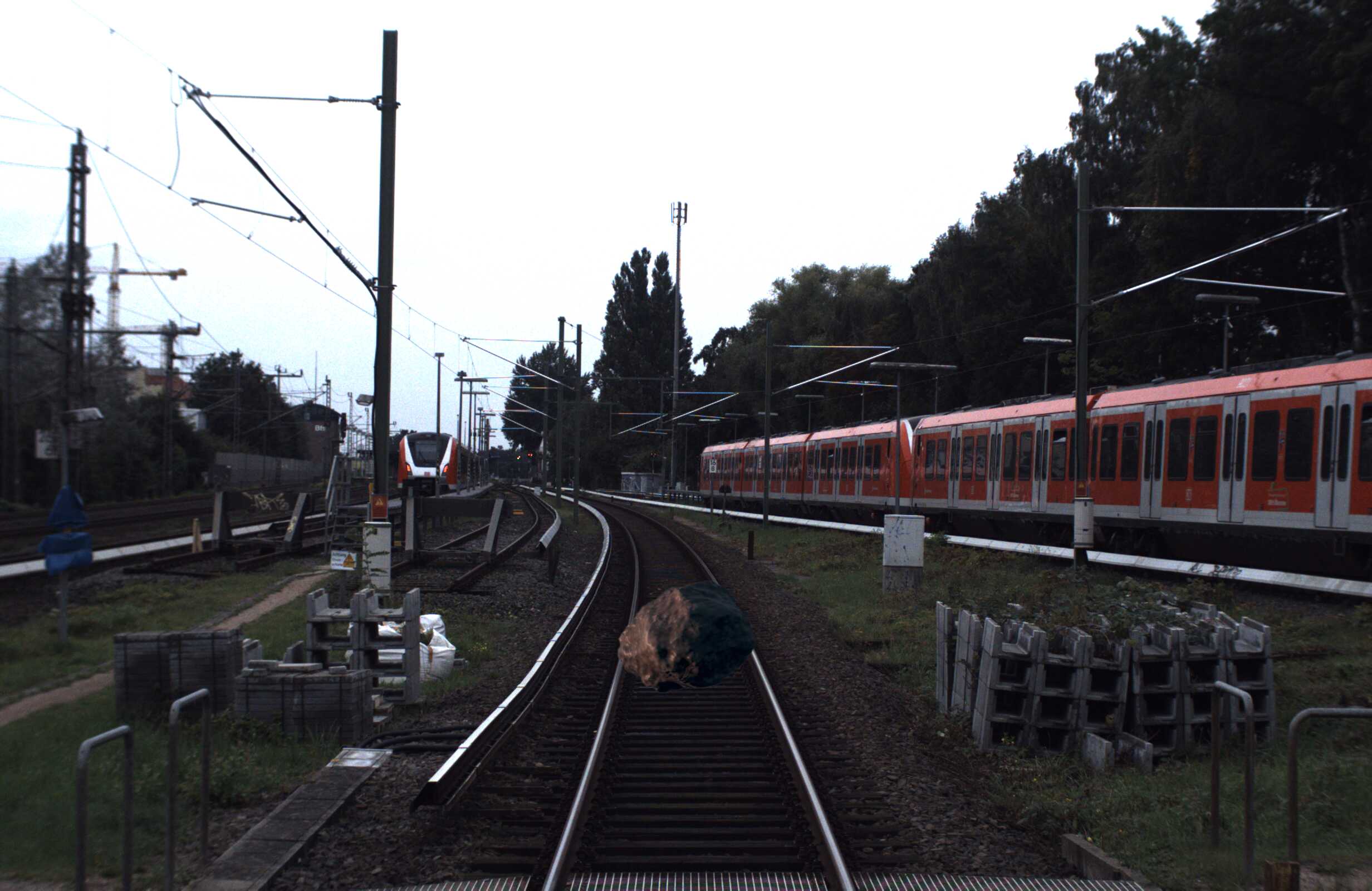}
        \\ \small (c)
    \end{minipage}
    \captionof{figure}{Three samples from the OSDaR-AR dataset: (a) a frame from sequence \seqthree~(\rgbhirescenter~camera) augmented with a virtual animal; (b) a point-cloud from sequence \seqsix~augmented with a virtual person; (c) a frame from sequence \seqfive~(\rgbcenter~camera) augmented with a virtual boulder. Best viewed in digital version. }
    \label{fig:teaser}
    \vspace{2mm}
\end{strip}

\iffalse
\twocolumn[{%
    \renewcommand\twocolumn[1][]{#1}%
    \maketitle

    \begin{center}
        \centering
        %\vspace{-5mm} 
        
        \begin{minipage}{0.33\linewidth}
            \centering
            %\includegraphics[trim={10 cm} {7 cm} {10 cm} {7 cm}, clip,width=\textwidth]{figures/vlcsnap-2026-02-05-11h57m02s537.png}
            \includegraphics[width=\textwidth]{figures/045_compressed.jpg}
            %\vspace{0.2pt}
            \\ \small (a)
        \end{minipage}
        \hfill
        \begin{minipage}{0.28\linewidth}
            \centering
            %\includegraphics[trim={10 cm} {7 cm} {10 cm} {7 cm}, clip,width=\textwidth]{figures/vlcsnap-2026-02-05-12h26m28s385.png}
            \includegraphics[width=\textwidth]{figures/pcd_sample2.png}
            %\vspace{0.2pt}
            \\ \small (b)
        \end{minipage}
        \hfill
        \begin{minipage}{0.33\linewidth}
            \centering
            \includegraphics[width=\textwidth]{figures/094_compressed.jpg}
            %\vspace{0.2pt}
            \\ \small (c)
        \end{minipage}

        %\vspace{4mm}
        \captionof{figure}{Three samples from the OSDaR-AR dataset: (a) a frame from sequence \seqthree~(\rgbhirescenter~camera) augmented with a virtual animal; (b) a point-cloud from sequence \seqsix~augmented with a virtual person; (c) a frame from sequence \seqfive~(\rgbcenter~camera) augmented with a virtual boulder. Best viewed in digital version. }
        \label{fig:teaser}
        %\vspace{5mm}
    \end{center}
}]
\fi
    % ---------------------------
	\thispagestyle{empty}
	\pagestyle{empty}
	
	%%%%%%%%%%%%%%%%%%%%%%%%%%%%%%%%%%%%%%%%%%%%%%%%%%%%%%%%%%%%%%%%%%
	\begin{abstract}
		% Replace with your abstract
		Although deep learning has significantly advanced the perception capabilities of intelligent transportation systems, railway applications continue to suffer from a scarcity of high-quality, annotated data for safety-critical tasks like obstacle detection. 
        While photorealistic simulators offer a solution, they often struggle with the ``sim-to-real" gap; conversely, simple image-masking techniques lack the spatio-temporal coherence required to obtain augmented single- and multi-frame scenes with the correct appearance and dimensions. 
        This paper introduces a multi-modal augmented reality framework designed to bridge this gap by integrating photorealistic virtual objects into real-world railway sequences from the OSDaR23 dataset. 
        Utilizing Unreal Engine 5 features, our pipeline leverages LiDAR point-clouds and INS/GNSS data to ensure accurate object placement and temporal stability across RGB frames. 
        This paper also proposes a segmentation-based refinement strategy for INS/GNSS data to significantly improve the realism of the augmented sequences,
        as confirmed by the comparative study presented in the paper. 
        Carefully designed augmented sequences are collected to produce OSDaR-AR, a public dataset designed to
        support the development of next-generation railway perception systems. The dataset is available at the following page: %\footnote{Reviewers may download a confidential preview at 
        %\textcolor{blue}
        \URL 
        %(copy-paste in a browser)}
        .
	\end{abstract}
	
	%%%%%%%%%%%%%%%%%%%%%%%%%%%%%%%%%%%%%%%%%%%%%%%%%%%%%%%%%%%%%%%%%%
	\input{01_intro}

	%%%%%%%%%%%%%%%%%%%%%%%%%%%%%%%%%%%%%%%%%%%%%%%%%%%%%%%%%%%%%%%%%%
	\input{02_related}

	%%%%%%%%%%%%%%%%%%%%%%%%%%%%%%%%%%%%%%%%%%%%%%%%%%%%%%%%%%%%%%%%%%
	\input{03_method}

    %%%%%%%%%%%%%%%%%%%%%%%%%%%%%%%%%%%%%%%%%%%%%%%%%%%%%%%%%%%%%%%%%%
	\input{04_exp}

    %%%%%%%%%%%%%%%%%%%%%%%%%%%%%%%%%%%%%%%%%%%%%%%%%%%%%%%%%%%%%%%%%%
	\input{05_dataset}

    %%%%%%%%%%%%%%%%%%%%%%%%%%%%%%%%%%%%%%%%%%%%%%%%%%%%%%%%%%%%%%%%%%

\input{06_conclusions}

	%%%%%%%%%%%%%%%%%%%%%%%%%%%%%%%%%%%%%%%%%%%%%%%%%%%%%%%%%%%%%%%%%%
	\section*{ACKNOWLEDGMENTS}
	This work has been partially supported by project SERICS (PE00000014) under the NRRP MUR program funded by the EU - NGEU.
	
	%%%%%%%%%%%%%%%%%%%%%%%%%%%%%%%%%%%%%%%%%%%%%%%%%%%%%%%%%%%%%%%%%%
	%\addtolength{\textheight}{-12cm}
	%\vspace{10mm}
    %\newpage
	\bibliographystyle{IEEEtran}
	% Your .bib file here
	\bibliography{root} 
	
\end{document}

%% file: 01_intro.tex
\section{Introduction}\label{s:intro}

% \begin{figure*}[t]
%     \centering
%     % Subfigure 1
%     \begin{subfigure}[b]{0.3\linewidth}
%         \centering
%         \includegraphics[width=\textwidth]{figures/vlcsnap-2026-02-05-11h57m02s537.png}
%         \caption{Description A}
%     \end{subfigure}
%     \hfill % Adds spacing between images
%     % Subfigure 2
%     \begin{subfigure}[b]{0.3\linewidth}
%         \centering
%         \includegraphics[width=\textwidth]{figures/vlcsnap-2026-02-05-11h57m02s537.png}
%         \caption{Description B}
%     \end{subfigure}
%     \hfill
%     % Subfigure 3
%     \begin{subfigure}[b]{0.3\linewidth}
%         \centering
%         \includegraphics[width=\textwidth]{figures/vlcsnap-2026-02-05-11h57m02s537.png}
%         \caption{Description C}
%     \end{subfigure}
    
%     \caption{Main caption for all three images.}
%     \label{fig:three_images}
% \end{figure*}

%\item Intro on AI in transportations systems and importance of data
Recently, artificial intelligence (AI) revolutionized the development and performance of intelligent transportation systems~\cite{veres2020survey}. Neural networks and deep learning significantly improved perception and decision making capabilities, becoming the de-facto standard for many applications.
The performance of a deep learning model is usually linked to the amount of data available for training, as well as the quality of data and the distribution consistency between training and testing datasets.  

%\item Intro on railway obstacle detection and lack of data
The amount of training data is crucial when considering deep learning applications for railway environments. In fact, AI perception for railway systems has fallen behind other more popular applications such as autonomous driving, despite employing similar technology. This lack of data is mainly due to the strict safety rules that apply to the operation of railway systems: data acquisition campaigns require the installation of dedicated hardware, the utilization of expensive vehicles, and the occupation of the infrastructure used for public transportation. On top of that, obtaining representative data for tasks such as obstacle or intrusion detection is dangerous and typically not permitted. 

%\item Intro on synthetic data and limitations (realism for virtual + temporal and visual coherence for stickers)
These hurdles motivated the community to develop photorealistic simulators for railway environments that are designed to generate datasets with automatic annotations. %The images generated with these simulators, even though realistic, usually do not fully capture real-world conditions and variability, resulting in sim-to-real transfer that often requires additional effort to use synthetic data for training and evaluation
The images generated with these simulators, while realistic, do not fully capture real-world conditions and variability. This results in sim-to-real transfer challenges that require additional effort when using synthetic data for training and evaluation~\cite{manivasagam2023towards}.

An alternative approach proposed in the literature is based on cropping objects from real-world images and placing them onto real-world images of railway environments~\cite{guo2025synrailobs, brucker2023local}. Although this approach is fast and effective, it does not provide accurate placement, scaling, or temporal coherence. 

Sitting at the intersection between these two methods, augmented reality (AR) takes the best of both worlds: starting from real-world images, it enhances them by adding objects rendered with a photorealistic simulator, enabling accurate placement and ensuring temporal coherence between the frames. However, despite being used in autonomous driving, AR was never employed to augment sequences captured for railway environment perception.

%intro on contributions.
Motivated by these factors, this paper presents a multi-modal augmented reality framework based on Unreal Engine 5 (UE5)~\cite{unrealengine5} and specifically designed for railway data sequences. 
The multi-modal data sequences are taken from the OSDaR23 dataset~\cite{tagiew2023osdar23}, which enable AR thanks to presence of LiDAR point-clouds and INS/GNSS data in addition to the RGB images. 
The rendering pipeline reconstructs and replicates a minimal version of the environment with simplified shapes and moves the virtual camera in the same position and orientation as in the real-world sequence. 
Early experimental results revealed that the available localization data in the OSDaR23 dataset (INS/GNSS) is, in some occasion, not properly aligned with the other sensors, 
making the position of the rendered objects not stable across frames. 
%This results in poor AR performance in terms of stable and realistic positioning of the objects. 
For this reason, this work also proposes a segmentation-based localization refinement 
%of the INS/GNSS positions 
that identifies the point-cloud portion belonging to the track and projects the INS/GNSS positions onto the centerline of the reconstructed track. This approach visibly and quantitatively improves the realism of the sequences by significantly reducing pixel reconstruction error and jitter.
As a result, this paper presents OSDaR-AR, a public dataset for multi-modal intrusion and obstacle detection in railway systems, consisting of 18 augmented sequences (for a total of 1800 frames). A few samples from the dataset are showed in Figure~\ref{fig:teaser}.
%contributions recap

To summarize, this paper introduces the following contributions:
\begin{itemize}
    \item it proposes an UE5-based pipeline for multi-modal augmented reality in railway environments (RGB camera and LiDAR point-cloud);
    \item it discusses a segmentation-based refinement method for INS/GNSS data;
    \item it performs a comparative study to identify the best localization strategy for AR applications in rail perception;
    \item it proposes OSDaR-AR, a public dataset with 18 AR-augmented OSDaR23 sequences that contain obstacles and their annotations. The dataset is available at the following link: \URL
\end{itemize}

The remainder of this paper is organized as follows: Section \ref{s:related} briefly reviews the related literature, Section \ref{s:method} introduces the proposed methods, Section \ref{s:exp} shows the experimental results obtained, Section \ref{s:dataset} illustrates the public dataset generated, and finally Section \ref{s:conclusions} discusses current limitations and future extensions.

%% file: 02_related.tex
\section{Background and related works}\label{s:related}

% \item Background on real-world data for railways. 
%One of the main factors that is limiting the development of deep learning perception solutions for railway environments is the lack of real-world data for training and extensive evaluation of the algorithms. 
%A notable exception is RailSem19~\cite{zendel19railsem}, a dataset with 8500 RGB and semantic segmentation images built from a wide variety of public YouTube videos and annotated with a semi-automatic process. 
%The dataset is general enough to sufficiently capture the typical distribution of railway environments. However, it only provides single camera RGB images and SS annotations, which limits the scope of the dataset.
%Another public dataset is OSDaR23~\cite{tagiew2023osdar23}, a multi-modal dataset for scene understanding. OSDaR23 includes images from multiple RGB and infrared cameras, as well as synchronized GNSS/INS localization and LiDAR point-clouds, making it suitable for extensive scientific investigation in the railway domain. 
%Unfortunately, only a few of the sequences include a moving train for more than 10 consecutive frames, therefore limiting the scope of the dataset.

% TODO insert RailGoerl24, RailSet (Rail-Ano), 
%Similarly to OSDaR23, RailGoerl24~\cite{tagiew2025railgoerl24} is a multi-modal dataset that includes longer sequences depicting people moving on the track. However, for safety reasons, the train had to travel the track in reverse and, in many sequences, the train is not moving at all.

A key bottleneck in developing deep learning perception for railway environments is the scarcity of real-world training and evaluation data. RailSem19~\cite{zendel19railsem} addresses this with 8500 RGB images and semantic segmentation annotations derived from YouTube videos, offering broad coverage of railway scenes but limited to single-camera RGB and segmentation only. OSDaR23~\cite{tagiew2023osdar23} provides richer multi-modal data—including RGB, infrared, LiDAR, and GNSS/INS—but contains few sequences with moving trains exceeding 10 frames. RailGoerl24~\cite{tagiew2025railgoerl24} similarly offers multi-modal sequences with on-track pedestrians, though safety constraints required reverse travel and stationary recording in many cases.

%Other public datasets are Gerald~\cite{leibner2023gerald} and FRSign~\cite{harb2020frsign}, which only include railway sign annotations.

% \item Background on synthetic data for railways. 

Due to the lack of real-world data, several works proposed the use of photorealistic simulators ~\cite{d2023trainsim} to generate synthetic datasets~\cite{d2025syndra, degordoa2023scenario, toprak2020conditional, diaz2025towards, broekman2021railenv, neri2022object}. 
Despite the adoption of realistic 3D graphic engines, these datasets suffer from sim-to-real gap and could not substitute real-world data completely~\cite{d2025syndra}.

% synrailobs, fishyrails
As an alternative, researchers created obstacle and anomaly detection datasets by cropping objects from image classification or object detection datasets, and pasting them onto the images coming from railway datasets~\cite{brucker2023local, guo2025synrailobs}. However, this approach is heavily limited for the following factors: (i) the appearance and scale of the pasted object is not realistic; (ii) the approach only works for single frames; and (iii) no point-cloud data can be reliably generated.

% \item Background on AR for ITS. 
Augmented reality techniques can provide an interesting solution for improving the imperfect realism of virtual railway scenes. Such techniques have been employed for several robotics and self-driving applications~\cite{argui2024advancements}, but, to our best knowledge, they were never employed for railway perception systems. AR may be extremely beneficial in railway environments, where both the lack of real-world data and safety issues make the acquisition of obstacle detection datasets a real challenge.  
In addition, the use of AR for railway applications has the advantage of improving the appearance, scale, and temporal coherence throughout the entire sequence, thus obtaining highly realistic multi-modal augmented sequences with respect to all the previous approaches.

% Review on localization refinement?

%% file: 03_method.tex
\section{Proposed Methods}\label{s:method}

\begin{figure*}
    \centering
    \includegraphics[width=.9\linewidth]{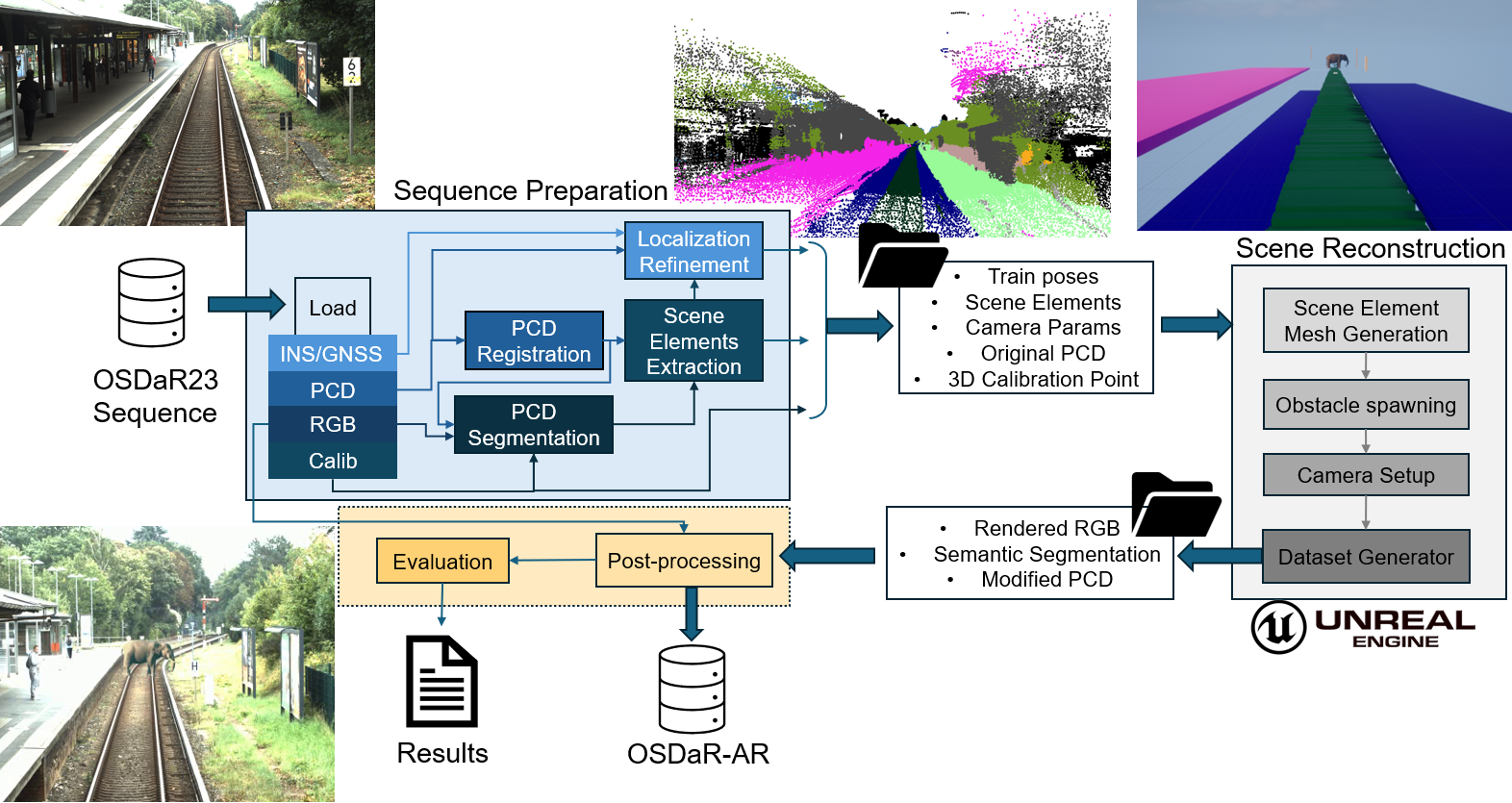}
    \caption{Preparation and rendering pipeline to obtain OSDaR-AR sequences from the original OSDaR data. The images also show, from top left to bottom left clock-wise, a sample RGB image from sequence \seqsix~(\rgbhirescenter~camera), the segmented point-cloud, the minimal reconstructed virtual environment, and the final rendered frame (cropped to better show details).}
    \label{fig:pipeline}
\end{figure*}

%This section provides an overview of the AR sequence preparation and rendering pipeline (Section \ref{s:method_overview}), then delves into the details of the Sequence Preparation (Section \ref{s:method_reconstruction}), the UE5-based scene reconstruction and rendering (Section \ref{s:method_rendering}), and the post-processing methods (Section \ref{s:method_post}). 

%\subsection{Pipeline Overview}\label{s:method_overview}

The proposed AR pipeline, illustrated in Figure~\ref{fig:pipeline}, is divided into three main phases:

\textbf{Sequence preparation.} The objective of this phase is to pre-process the multi-modal sequence of data from the OSDaR23 dataset to obtain distilled information necessary to reconstruct a minimal digital twin of the environment where to place the virtual objects and cameras in the correct positions. In this phase, an accurate pose estimation of the real cameras is crucial to correctly place the corresponding virtual cameras in the synthetic environment and obtain a realistic AR sequence. 
Additionally, it is also important to obtain the position and shape of fundamental scene elements, such as rails, platforms, and poles among others. 
%This is done by registering a complete point-cloud by joining the points coming from subsequent frames, then obtaining semantic information about each point through the segmentation of RGB images. 
These aspects are described in detail in Section \ref{s:method_reconstruction}.

\textbf{Virtual scene reconstruction.}
This phase, performed exclusively in UE5, is dedicated to the virtual scene reconstruction, which is important to place obstacles coherently to the scene layout. Then, the cameras are placed in the environment with the correct intrinsic and extrinsic parameters; finally, the virtual RGB images and point-clouds are rendered and saved together with the segmentation mask of the rendered virtual obstacles. This phase is described in detail in Section \ref{s:method_rendering}.

\textbf{Post-processing and evaluation.}
The raw data from UE5 must be post-processed to match the illumination conditions of the original sequence, add motion blurring effects, and combine original and virtual elements. This phase, which may also include a quantitative evaluation of the quality of the AR setup, is detailed in Section \ref{s:method_post}.

\subsection{Sequence preparation}\label{s:method_reconstruction}
The objective of this stage is to pre-process the original OSDaR data sequence to obtain the information required by the graphics engine to accurately setup the synthetic image generation. In particular, the information required are (i) the train poses throughout the sequence, (ii) the intrinsic and extrinsic camera parameters, (iii) the original point-cloud, and (iv) a set of scene elements useful for virtual obstacle placement. Additionally, it is useful to provide the position of a 3D calibration point for quantitatively evaluate the goodness of the AR solution. This particular aspect is better explained in Section \ref{s:method_post}.

% Load
OSDaR23 data are first loaded to obtain a sequence of INS/GNSS poses, the point-clouds, RGB images (a sample in the top-left corner of Figure \ref{fig:pipeline}), and the camera-LiDAR calibration matrices. 
% PCD registration
The first step is to register the point-clouds into a single point-cloud. This is useful to capture details of the whole scene from multiple points of view. 

\begin{figure}[ht]
    \centering
    % Row (a)
    %(a) \includegraphics[align=c, width=0.85\columnwidth]{figures/061_1631704655.200000011.png}
    
    %\vspace{1em} % Vertical gap between rows
    
    % Row (b)
    %(b) \includegraphics[align=c, width=0.85\columnwidth]{figures/PCDSegmentation.png}
    
    %\vspace{1em}
    
    % Row (c)
    %(c) \includegraphics[align=c, width=0.85\columnwidth]{figures/MinimalReconstruction.png}
    
    %\vspace{1em}
    
    % Row (d)
    \includegraphics[align=c, width=0.85\columnwidth]{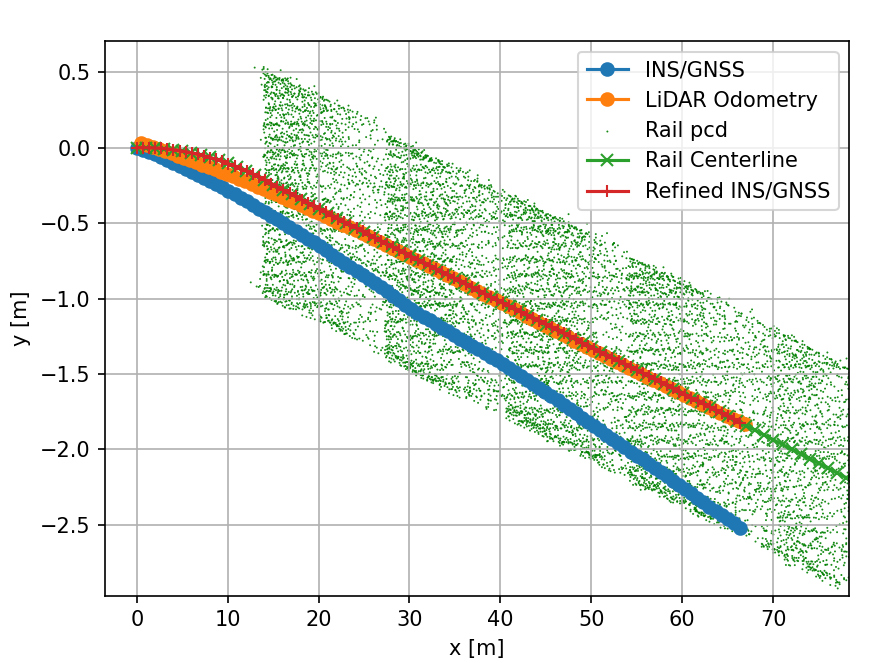}
    
    \caption{A comparison between different localization methods for sequence \seqsix: the INS/GNSS deviates from the rail geometry, the LiDAR Odometry is quite accurate, while the segmentation-based projection refinement of the INS/GNSS improves the original position.
    }
    \label{fig:main_figure}
\end{figure}

% scene segmentation 
The registered point-cloud is then segmented by using a semantic segmentation network specifically trained for railway environments. Namely, the DDRNet-Slim23~\cite{pan2022deep} was trained on RailSem19~\cite{zendel19railsem} as proposed by D'Amico et al.~\cite{d2025syndra}. The network is used to process the RGB images; the point-cloud is projected onto the images to identify the semantic class (top-central image in Figure \ref{fig:pipeline}). 
% scene element extraction (rail, ground-areas, poles, fences, etc)
The segmented point-cloud can be further processed to obtain the geometry of the rail track, which is useful %for precise virtual obstacle placement. 
% refined localization
%The rail geometry is also useful 
to refine the INS/GNSS data, which, in the OSDaR23 sequences, seem to be diverging from the actual train path. The INS/GNSS sequence is corrected by projecting the points onto the identified rail geometry. A visual comparison of these approaches is reported in Figure~\ref{fig:main_figure}, while quantitative comparison between INS/GNSS, LiDAR Odometry, and segmentation-based localization refinement is presented in Section~\ref{s:exp}.

% other objects
From the segmented point-cloud it is possible to also extract other elements. The areas next to the rail track and platforms are also useful for accurate virtual object placement. Fitting a plane with RANSAC~\cite{martinez2022ransac} on the point-cloud portions is enough to obtain a simplified representation of the environment. Similarly, individual poles can be identified by clustering~\cite{schubert2017dbscan}.

\subsection{Virtual scene reconstruction}\label{s:method_rendering}
The objective of the second phase is to reconstruct a minimal virtual replica of the real-world environment. %This enables accurate obstacle placement in specific portions of the scene. 
To do so, virtual cameras are set up with correct intrinsic and extrinsic parameters to mimic the real-world point of view of the environment.
This part of the pipeline is completely coded and executed in UE5. 

% Scene element mesh generation
First, scene element meshes are generated procedurally and spawned. Starting from the segmented points extracted in the first phase, it is quite straightforward to build plane-like meshes for platforms and areas close to the track. Poles are spawned as cubes stretched in the z-axis. The rail track itself is generated as a set of box-like meshes with fixed width and height that are spawned wherever points with track semantic label are present, following the real track geometry.

% obstacle/calibration spehere spawning
Thanks to this minimal virtual reconstruction of the environment, it is possible to spawn objects in the UE5 editor in the desired places, for instance, on the track or on the near areas, or on the edge of the platform. The minimal reconstructed environment is visualized in the top-right image of Figure \ref{fig:pipeline}.
In this phase, if available, the 3D calibration point can be spawned as a sphere. Details are provided in Section \ref{s:method_post}.

% camera (custom stencil and rendering)
Two cameras (\texttt{SceneCaptureComponents} in UE5) are required for AR: the first captures RGB images, while the second captures UE's \texttt{CustomStencil}, which corresponds to semantic segmentation. It is crucial to always place the two cameras in the same exact position with the same orientation, but also assign the same \texttt{CustomStencil} value to all the objects that must be rendered in the final scene. In fact, the resulting segmentation is a binary mask, accurately localizing the position of the object in the image. 
This information will be used in post-processing to combine the original images with the synthetic ones cropped according to the corresponding segmentation masks. 

% rendering, dataset generation
The two cameras are positioned according to the localization and calibration data obtained from the Sequence preparation phase. For each pose, the two cameras are moved and frames are rendered and collected. 

% PCD augmentations. 
In this phase, the point-cloud data are also augmented. For each frame, the original points are loaded and raycasting is used to find the points that are occluded by a virtual object with respect to the LiDAR sensor frame origin. The modified point-cloud is saved in a dedicated folder.
% insert comment on accuracy of this method and justifications. 
A limitation of our data augmentation pipeline is the assumption of a single raycasting origin. In fact, since the OSDAR23 point clouds are generated by merging inputs from six separate LiDAR sensors—whose individual poses are unavailable—we cannot perform exact occlusion culling for each distinct sensor. However, we consider this approximation valid for our specific use case. Since our virtual obstacles are placed centrally and in the immediate frontal proximity of the train, the simulated occlusion primarily affects the frontal sensor's field of view, thereby preserving the geometric consistency of the most relevant data points.

\subsection{Post-processing and evaluation}\label{s:method_post}
The third and last phase is designed to post-process the generated dataset to obtain more realistic AR, and to optionally evaluate the goodness of the augmented sequence.

% post-processing (blurring, illumination matching,
Before augmenting the original images with the cropped objects from the synthetic ones, it is important to match the illumination statistics between the two domains. With a simple image conversion to the LAB color space, we collect lightness statistics from the real-world sequence and modify the synthetic images to match it. 

Blurring is another important aspect to enhance realism. Although the real-world blurring heavily depends on exposure time, lighting conditions, and the train speed, we found that, for this application, simple Gaussian blur is enough to make the virtual objects blend in the real-world ones. The blurring kernel dimension is set depending on the distance to the object in the specific frame. 

%calibration evaluation + reprojection error and jitter)
Another important aspect is the evaluation of the quality of the AR solution in terms of the stability of the virtual objects in the scene. The accuracy of the localization directly impacts on the virtual camera point-of-view. The closest the real-world and virtual camera poses are, the better the perceived quality of the resulting augmented sequence will be. 

However, it might be tricky to quantify the accuracy of the localization method, as there is no ground-truth to compare against. To this end, for each OSDaR23 sequence, the image coordinates corresponding to a feature that could be easily recognized (e.g., corners of signs and top of poles) and that also appear in the point-cloud were manually collected.
This 3D calibration point is chosen by selecting the point in the point-cloud that, once projected, minimizes the pixel distance to the annotated coordinates. 
Once the 3D calibration point corresponding to the ground-truth coordinates is found, it can be used to evaluate the stability of the AR solution by placing a calibration sphere in the virtual environment (as mentioned in the previous section), and compute the reprojection error of the rendered sphere with respect to the manually collected coordinates. 
The reprojection error computed in this way can be used also to compensate the localization error by offsetting the syntetic image before merging it with the original one.
Additional details on the metrics can be found in Section \ref{s:exp}.

%To be more precise, it is best practice to consider the INS/GNSS data as ground-truth. However, we noticed that raw INS/GNSS is not suitable 

%% file: 04_exp.tex
\section{Experimental Results}\label{s:exp}
This section 
%provides details about the experimental setup employed in these tests. 
%Then, a comparison between different localization strategies is provided 
presents a set of experiments carried out 
to compare different localization strategies
in terms of AR solution quality.
%The localization methods compared are the raw INS/GNSS, LiDAR Odometry (KISS-ICP~\cite{vizzo2023kiss}), and our proposed segmentation-refined projection described in Section~\ref{s:method_reconstruction}.
In particular, the following localization strategies are compared:
\begin{itemize}
    \item INS/GNSS: This consists of the raw localization data originally included in the OSDaR23 dataset, obtained through inertial navigation and GNSS data fusion.
    \item LiDAR Odometry: This comprises the poses obtained by processing the LiDAR sequences with the KISS-ICP method~\cite{vizzo2023kiss}, which was selected for its superior performance compared to other LiDAR odometry methods. 
    \item Segmentation-refined: This refers to our proposed segmentation-refined projection method, as detailed in Section~\ref{s:method_reconstruction}.
\end{itemize}

\subsection{Experimental setup}
% HW and SW
The tests reported in this section were obtained on a Windows 11 PC with i9-9900 CPU @ 3.10GHz, 32GB RAM, and NVIDIA GeForce RTX 4060 Ti. UE5.2 and python3.10 were employed.   

% sequences analyzed
The OSDaR23 sequences used for the comparison are \seqthree, \seqfive, and \seqsix, which are the only three sequences that include a moving train for 100 frames. The other sequences were not considered, as the train is barely moving and only 10 frames were collected. For each experiment in this study, the sensors used are the LiDAR and a frontal camera. 
Considering that the \osdar~dataset includes two frontal RGB cameras (namely \rgbcenter~and \rgbhirescenter), the experiments were repeated for each camera.

% metrics: reprojection errors, jitter.
The metrics used for this comparison are the reprojection pixel error (RPE) and the positioning jitter, which are typical metrics for the quantitative evaluation of the goodness of AR solutions~\cite{louis2019real, azuma1997survey, holloway1997registration}. 
The RPE is an absolute measure in pixels that provides an estimate of the distance between the virtual and real camera poses. 
Jitter is a measure of the stability of the AR solution across consecutive frames.
Since the scene augmentation is performed offline, the processing time, typically crucial in AR pipelines, is not particularly relevant and is not considered in the comparison.
%In this offline setup where real-time performance and latency is not a particular issue, jitter is the key metric linked to the human realism perception~\cite{holloway1997registration}. 
In each frame $i$, the RPE is defined as $RPE_i = \|\mathbf{GT}_i-\mathbf{p}_i\|$, where $\mathbf{GT}_i$ corresponds to the manually collected ground-truth coordinates defined in Section \ref{s:method_post}, and $\mathbf{p}_i$ are the coordinates of the rendered 3D calibration sphere.  At each frame $i$, jitter is computed as $J_i = |RPE_{i+1}-RPE_i|$.

\subsection{Results}
%\begin{itemize}
    %\item Comparative study on best localization strategy for AR in railways.
%\end{itemize}

%able \ref{t:calib_results} reports the accuracy results for each sequence organized by camera source. 

For each sequence, the mean and standard deviation of the RPE and jitter were computed for each localization strategy. The achieved results are reported in Table \ref{t:calib_results}. 
%Considering that the \osdar~dataset includes two frontal RGB cameras (namely \rgbcenter~and \rgbhirescenter), the experiment was repeated for each camera. Clearly, a dedicated ground-truth must be collected for each camera sequence, as described in Section \ref{s:method_post}.

\input{calib_table}

As anticipated in Section \ref{s:method_reconstruction} and illustrated in Figure~\ref{fig:main_figure}, the results confirm that the raw INS/GNSS data in the OSDaR23 sequences are not particularly accurate and therefore not reliable for AR applications. Conversely, LiDAR Odometry and the segmentation-refined projection behave similarly, with a slight predominance of the latter, depending on the sequence, the camera used, and, above all, the accuracy of the manual ground-truth annotation. 
Specifically, it is not straightforward to provide stable and accurate annotations, as the world point selected as a reference must (i) stand out visually to simplify the annotation process, (ii) be visible and not occluded for the entire sequence, and (iii) correspond to a point in the 3D point-cloud. 
These aspects directly impact this kind of quantitative evaluation, hence it is not trivial to discriminate the best method for such AR solutions.
Nonetheless, a few key aspects emerge. 
First, the raw INS/GNSS is never the best choice: only in one case it produced the best jitter (comparable with the other methods), but with a worse reprojection error. The bad performance of the raw INS/GNSS could be explained with a sensor misalignment between the GNSS antenna and the camera-LiDAR setup; however, it seems that there is no constant misalignment between the sequences, and the real reasons for this should be investigated together with the OSDaR23 hardware setup. 
Second, the proposed segmentation-refined solution improves the raw INS/GNSS in almost all the sequences, especially in terms of reprojection error. When jitter is not improved, it has comparable values. 
Third, the LiDAR Odometry almost always provides reliable results, especially in terms of jitter. 

In summary, the LiDAR Odometry and our proposed segmentation-refined method show comparable performance, but a definitive winner could not be picked in this context, as additional testing must be carried out on additional sequences (which are scarce), and including multiple calibration points in each sequence. However, 
performing a comprehensive experimental champaign is out of the scope of this paper, which is mainly focused on proposing the AR pipeline and the OSDaR-AR dataset. Therefore, more specific experiments aimed at understanding the problems outlined above are left as a future work.
%since the main contribution of this work is the AR pipeline and the OSDaR-AR dataset, considering a single winner is not crucial: it is enough to choose the best-performing method for each sequence to generate the corresponding augmented sequences.  

% TODO a final remark on execution times?

%% file: calib_table.tex
\begin{table}[ht]
    \centering
    \footnotesize
    \setlength{\tabcolsep}{3pt} 
    \caption{Experimental results. For each Camera (C1: \rgbcenter, C2: \rgbhirescenter), top: $RPE$ [px], bottom: $J$ [px]. The best result in each row is highlighed in \textbf{bold}.}
    \begin{tabular}{lll ccc}
        \toprule
        & & & \makecell{INS/\\GNSS} & \makecell{LiDAR\\Odometry} & \makecell{Segmentation-\\refined (ours)} \\
        \midrule
        \multirow{4}{*}{\textbf{Seq 3.1}} & \multirow{2}{*}{\scriptsize C1} & 
        %3.1 - rgbcenter - rpe
        $RPE$ & $81.3 \pm 50.1$ & $\mathbf{25.3 \pm 17.4}$ & $31.1 \pm 12.5$ \\
        %3.1 - rgbcenter - j
        & & $J$ & $2.71 \pm 3.16$ & $1.91 \pm 2.70$ & $\mathbf{1.54 \pm 1.62}$ \\
        \cmidrule{2-6}
        %3.1 - rgbhighrescenter - rpe
        & \multirow{2}{*}{\scriptsize C2} & $RPE$ & $89.0 \pm 92.0$ & $56.8\pm 45.1$ & $\mathbf{22.0 \pm 27.7}$ \\
        %3.1 - rgbhirescenter - j
        & & $J$ & $4.23\pm 5.59$ & $3.11\pm 4.31$ & $\mathbf{2.14\pm 2.46}$ \\
        \midrule[\heavyrulewidth] % Thicker line between sequences
        %5.1 - rgbcenter - rpe
        \multirow{4}{*}{\textbf{Seq 5.1}} & \multirow{2}{*}{\scriptsize C1} & $RPE$ & $63.7\pm 44.7$ & $\mathbf{43.9\pm 27.0}$ & $60.1 \pm 43.5$ \\
        %5.1 - rgbcenter - j
        & & $J$ & $3.10 \pm 4.07$ & $\mathbf{2.17 \pm 2.43}$ & $3.10 \pm 4.03$ \\
        \cmidrule{2-6}
        %5.1 - rgbhirescenter - rpe
        & \multirow{2}{*}{\scriptsize C2} & $RPE$ & $77.9\pm 106.6$ & $92.5\pm 70.8$ & $\mathbf{77.0 \pm 107.7}$ \\
        %5.1 - rgbhighrescenter - j
        & & $J$ & $5.89 \pm 7.44$ & $\mathbf{4.43 \pm 5.55}$ & $5.92 \pm 7.56$ \\
        \midrule[\heavyrulewidth] % Thicker line between sequences
        %6.2 - rgbcenter - rpe
        \multirow{4}{*}{\textbf{Seq 6.2}} & \multirow{2}{*}{\scriptsize C1} & $RPE$ & $49.3 \pm 13.0$ & $43.6 \pm 5.3$ & $\mathbf{42.0 \pm 9.3}$ \\
        %6.2 - rgbcenter - j
        & & $J$ & $\mathbf{1.23 \pm 0.99}$ & $1.29 \pm 0.96$ & $1.26 \pm 0.93$ \\
        \cmidrule{2-6}
        %6.2 - rgbhighrescenter - rpe
        & \multirow{2}{*}{\scriptsize C2} & $RPE$ & $31.4 \pm 15.1$ & $20.9 \pm 4.9$ & $\mathbf{18.9 \pm 9.2}$ \\
        %6.2 - rgbhighrescenter - j
        & & $J$ & $1.45 \pm 1.32$ & $\mathbf{1.29 \pm 1.22}$ & $1.78 \pm 1.61$ \\
        \midrule[\heavyrulewidth] % Thicker line between sequences
        \bottomrule
    \end{tabular}
    \label{t:calib_results}
\end{table}
%\vspace{-5pt}

%% file: 05_dataset.tex
\section{OSDaR-AR dataset}\label{s:dataset}
The proposed OSDaR-AR dataset was generated with the pipeline described in Section \ref{s:method}. It comprises 18 multi-modal sequences, obtained by spawning 6 different virtual objects (i.e., \textit{person}, \textit{rock}, \textit{fallen tree}, \textit{cow}, \textit{horse}, \textit{indian elephant}) in each of the three 100-frames-long sequences available in OSDaR23 (namely, \seqthree, \seqfive, and \seqsix), for a total of 1800 frames.
For each sequence, data from the \rgbcenter~and the \rgbhirescenter~cameras were augmented, as well as the point-clouds from the LiDAR sensor. Hence, the dataset does not replicate the OSDaR23 sensor suite completely; rather, the sensors that were considered in this work are the ones that are mostly used for obstacle or anomaly detection applications. The data is saved in the same formats used in the original OSDaR23 dataset.
Future extensions will also address additional sensors, such as lateral cameras, infrared cameras, and radar data. 
% TODO add comment on additional sequences. 
Additional sequences will be added in the future to extend the dataset with additional real-world scenarios and/or virtual objects, also depending on the feedbacks collected from the research community and industrial stakeholders. 
For space limitations, only a few samples from the sequences are showed in Figure~\ref{fig:teaser}. The spatio-temporal coherence throughout the sequences is better appreciated in video format at the project webpage: \URL.  
%Graphical assets were sourced from \textcolor{blue}{TODO INSERT LINKS.}

%% file: 06_conclusions.tex
\section{Conclusions}\label{s:conclusions}
This paper presented a multi-modal offline AR framework based on Unreal Engine 5 specifically designed for railway environments. 
The framework loads OSDaR23 sequences and processes them to reconstruct a minimal virtual environment, which in turn enables accurate virtual object placement and rendering.
Since AR solutions are sensitive to localization errors, the raw localization data provided in OSDaR23 is refined with a segmentation-based approach, which improves the quality and realism of the augmented sequences. 
The generated sequences are collected and published as a public dataset, OSDaR-AR, which includes 18 augmented sequences that include two cameras and a LiDAR sensor.

% TODO list some limitations and future work?